\documentclass[preprint,12pt]{elsarticle}
\usepackage{amssymb}
\usepackage{helvet}
\usepackage{courier}
\usepackage{graphicx}
\usepackage{ifthen}
\usepackage{amsmath, amssymb}

\usepackage{eufrak}
\usepackage {float}
\usepackage[amsmath,thmmarks]{ntheorem}
\usepackage{mathrsfs}
\usepackage{verbatim}
\usepackage{array,color}
\usepackage[table]{xcolor}
\usepackage{url}

\def\pref{\vartriangleright}
\def\prefe{\trianglerighteq}

\newcommand{\ra}{\rightarrow}

\newcommand{\x}{{ \bf x}}

\newcommand{\mo}{{\mathcal O}}

\newcommand{\mc}{\mathcal C}

\newcommand{\gs}{\text{GS}}
\newcommand{\pr}[1]{\text{Pr}\left(#1\right)}
\newcommand{\order}{\text{Ord}}

\newcommand{\vo}{\text{\sc VO}}
\newcommand{\cvo}{\text{\sc CVO}}
\newcommand{\dvo}{\text{\sc DVO}}
\newcommand{\cwvo}{\text{\sc CWVO}}

\newcommand{\mcp}{{\cal P}}

\newcommand{\Omit}[1]{}
\newcommand{\fjerome}[1]{}

\newtheorem{thm}{Theorem}
\newtheorem{lemma}{Lemma}

\newtheorem{coro}{Corollary}

\newtheorem{claim}{Claim}

\newtheorem{dfn}{Definition}
\newtheorem{ex}{Example}

\newenvironment{prf}[2][t]{\noindent{\bf {Proof of \ifthenelse{\equal{#1}{t}}{Theorem}
{\ifthenelse {\equal{#1}{l}}{Lemma}{\ifthenelse
{\equal{#1}{c}}{Claim}{\ifthenelse
{\equal{#1}{coro}}{Corollary}{Proposition}}}} \ref{#2}:}}\ }{\hfill
$\blacksquare$ \vspace{2mm}}

\journal{Artificial Intelligence}

\begin{document}

\begin{frontmatter}
\title{How Many Vote Operations Are Needed to Manipulate a Voting System?}

\author[harvard]{Lirong Xia\corref{a}}
\ead{lxia@seas.harvard.edu}
\cortext[a]{Corresponding author. Tel:+1-617-495-1246.}
\address[harvard]{School of Engineering and Applied Sciences, Harvard University, Cambridge, MA 02138, USA}

\date{}

\begin{abstract}
In this paper, we propose a framework to study a general class of strategic behavior in voting, which we call {\em vote operations}. We prove the following theorem: if we fix the number of alternatives, generate $n$ votes i.i.d.~according to a distribution $\pi$, and let $n$ go to infinity, then for any $\epsilon >0$, with probability at least $1-\epsilon$, the minimum number of operations that are needed for the strategic individual to achieve her goal falls into one of the following four categories:
(1) $0$, (2) $\Theta(\sqrt n)$, (3) $\Theta(n)$, and (4) $\infty$. This theorem holds for any set of vote operations, any individual vote distribution $\pi$, and any {\em integer generalized scoring rule}, which includes (but is not limited to) almost all commonly studied voting rules, e.g., approval voting, all positional scoring rules (including Borda, plurality, and veto), plurality with runoff, Bucklin, Copeland, maximin, STV, and ranked pairs.

We also show that many well-studied types of strategic behavior fall under our framework, including (but not limited to) constructive/destructive manipulation, bribery, and control by adding/deleting votes, margin of victory, and minimum manipulation coalition size. Therefore, our main theorem naturally applies to these problems.
\end{abstract}


\end{frontmatter}

\section{Introduction}
Voting is a popular method used to aggregate voters' preferences to make a joint decision. Recently, voting has been used in many fields of artificial intelligence, for example in multi-agent systems~\cite{Ephrati91:Clarke}, recommender systems~\cite{Ghosh99:Voting,Pennock00:Social}, and web-search engines~\cite{Dwork01:Rank}. One of the most desired properties for voting rules is {\em strategy-proofness}, that is, no voter has incentive to misreport her preferences to obtain a better outcome of the election. Unfortunately, strategy-proofness is not compatible with some other desired properties, due to the celebrated Gibbard-Satterthwaite theorem~\cite{Gibbard73:Manipulation,Satterthwaite75:Strategy}, which states that when there are at least three alternatives, no strategy-proof voting rule satisfies the following two natural properties: non-imposition (every alternative can win) and non-dictatorship (no voter is a {\em dictator}, whose top ranked alternative is always selected to be the winner). 

Even though manipulation is inevitable, researchers have set out to investigate whether computational complexity can serve as a barrier against various types of strategic behavior, including manipulation. The idea is, if we can prove that it is computationally too costly for a strategic individual to find a beneficial operation, she may give up doing so. Initiated by Bartholdi, Tovey, and Trick~\cite{Bartholdi89:Computational}, a fair amount of work has been done to characterize the computational complexity of various types of strategic behavior\footnote{See~\cite{Faliszewski10:AI,Faliszewski10:Using,Rothe12:Typical} for recent surveys.}, including the following.

$\bullet$ {\em Manipulation}: a voter or a coalition of voters cast false vote(s) to change the winner (and the new winner is more preferred).

$\bullet$  {\em Bribery}: a strategic individual changes some votes by bribing the voters to make the winner preferable to her~\cite{Faliszewski09:How}. The bribery
problem is closely related to the problem of computing the {\em margin of
  victory}~\cite{Cary11:Estimating,Magrino11:Computing,Xia12:Computing}.

$\bullet$  {\em Control}: a strategic individual adds or deletes votes to make the winner more preferable to her~\cite{Bartholdi92:How}. 

Most previous results studying ``using computational complexity as a barrier against strategic behavior'' conduct worst-case analyses of computational complexity. Recently, an increasing number of results show that manipulation, as a particular type of strategic behavior, is typically not hard to compute. One direction, mainly pursued in the theoretical computer science community, is to obtain a quantitative version of the Gibbard-Satterthwaite theorem, showing that for any given voting rule that is ``far'' enough from any dictatorships, an instance of manipulation can be found easily with high probability. This line of research was initiated by Friedgut, Kalai, and Nisan~\cite{Friedgut08:Elections}, where they proved the theorem for $3$ alternatives and neutral voting rules. The theorem was extended to an arbitrary number of alternatives by Isaksson, Kindler, and Mossel~\cite{Isaksson10:Geometry}, and finally, the neutrality constraint was removed by Mossel and Racz~\cite{Mossel12:quantitative}. Other extensions include Dobzinski and Procaccia~\cite{Dobzinski08:Frequent} and Xia and Conitzer~\cite{Xia08:Sufficient}. 

Another line of research is to characterize the ``frequency of manipulability'', defined as the probability for a randomly generated preference-profile to be manipulable by a group of manipulators, where the non-manipulators' votes are generated i.i.d.~according to some distribution (for example, the uniform distribution over all possible types of preferences). Peleg~\cite{Peleg79:Note}, Baharad and Neeman~\cite{Baharad02:Asymptotic}, and Slinko~\cite{Slinko02:Asymptotic,Slinko04:How} studied the asymptotic frequency of manipulability for positional scoring rules when the non-manipulators' votes are drawn i.i.d.~uniformly at random. Procaccia and Rosenschein~\cite{Procaccia07:Average} showed that for positional scoring rules, when the non-manipulators’ votes are drawn i.i.d. according to some distribution that satisfies some natural conditions, if the number of manipulators is $o(\sqrt n)$, where $n$ is the number of non-manipulators, then the probability that the manipulators can succeed goes to $0$ as $n$ goes to infinity; if the number of manipulator is $\omega(n)$, then the probability that the manipulators can succeed goes to $1$.

This dichotomy theorem was generalized to a class of voting rules called {\em generalized scoring rules (GSRs)} by Xia and Conitzer~\cite{Xia08:Generalized}. A GSR is defined by two functions $f,g$, where $f$ maps each vote to a vector in multidimensional space, called a {\em generalized scoring vector} (the dimensionality of the space is not necessarily the same as the number of alternatives). Given a profile $P$, let {\em total generalized scoring vector} be the sum of $f(V)$ for all votes $V$  in $P$. Then, $g$ selects the winner based on the total preorder of the components of the total generalized scoring vector. We call a GSR an {\em integer}  GSR, if the components of all generalized scoring vectors are integers. (Integer) GSRs are a general class of voting rules. One evidence is that many commonly studied voting rules are integer GSRs, including (but not limited to) approval voting, all positional scoring rules (which include Borda, plurality, and veto), plurality with runoff, Bucklin, Copeland, maximin, STV, and ranked pairs.\footnote{
The definition of these commonly studied voting rules can be found in, e.g.,~\cite{Xia08:Generalized}. In this paper, we define GSRs as voting rules where the inputs are profiles of linear orders. GSRs can be easily generalized to include other types of voting rules where the inputs are not necessarily linear orders, for example, approval voting.} As another evidence, GSRs admit a natural axiomatic characterization~\cite{Xia09:Finite}, which also suggests that GSRs are equivalent to {\em hyperplane rules}~\cite{Mossel12:Smooth}. The knife-edge case of $\Theta(\sqrt n)$ was studied experimentally for STV and veto in~\cite{Walsh09:Where}, showing that the probability for the manipulators to succeed has a smooth phase transition. More recently, \cite{Mossel12:Smooth} extends the dichotomy theorem to all {\em anonymous} voting rules for distributions that satisfy some mild conditions, and theoretically proved that for all generalized scoring rules, for the knife-edge case, the probability that the manipulators can achieve their goal is continuously differentiable, which suggests a smooth phase transition.

While most of the aforementioned results are about manipulation, in this paper, we focus the optimization variants of various types of strategic behavior, including manipulation, bribery, and control. Despite being natural, to the best of our knowledge, such optimization variants have been investigated for only three types of strategic behavior. The first is  the {\em unweighted coalitional optimization (UCO)} problem, where we are asked to compute the minimum number of manipulators who can make a given alternative win~\cite{Zuckerman09:Algorithms}.  Approximation algorithms have been proposed for {\sc UCO} for specific voting systems, including positional scoring rules and maximin~\cite{Zuckerman09:Algorithms,Xia10:Scheduling,Zuckerman11:Algorithm}. 
The second is the {\em margin of victory} problem, where we are asked to compute the smallest number of voters who can change their votes to change the winner~\cite{Magrino11:Computing,Cary11:Estimating,Xia12:Computing}. The third is the {\em minimum manipulation coalition size} problem, which is similar to the margin of victory, except that all voters who change their votes must prefer the new winner to the old winner~\cite{Pritchard09:Asymptotics}.


\subsection{Our Contributions}
In this paper, we introduce a unified framework to study a class of strategic behavior for generalized scoring rules, which we call {\em vote operations}. In our framework, a strategic individual seeks to change the winner by applying some operations, which are modeled as vectors in a multidimensional space. We study three goals of the strategic individual: (1) making a favored alternative win, called {\em constructive vote operation ($\cvo$}), (2)  making a disfavored alternative lose, called {\em destructive vote operation ($\dvo$)}, and (3) change the winner of the election, called {\em change-winner vote operation ($\cwvo$)}.  The framework will be formally defined in Section~\ref{sec:model}. This is our main conceptual contribution.

Our main technical contribution is the following asymptotical characterization of the minimum number of operations that are needed for the strategic individual to achieve her goal.

\vspace{3mm}
\noindent {\bf Theorem~\ref{thm:main} (informally put)}
Fix the number of alternatives and the set of vote operations. For any integer generalized scoring rule and any distribution $\pi$ over votes, we generate $n$ votes i.i.d.~according to $\pi$ and let $n$ go to infinity. Then, for any $\vo\in\{\cvo,\dvo,\cwvo\}$ and any $\epsilon >0$, with probability at least $1-\epsilon$, the minimum number of operations that are necessary for the strategic individual to achieve $\vo$ falls into one of the following four categories:
(1) $0$, (2) $\Theta(\sqrt n)$, (3) $\Theta(n)$, and (4) $\infty$.
\vspace{3mm}

More informally, Theorem~\ref{thm:main} states that in large elections, to achieve a specific goal (one of the three goals described above), with probability that can be infinitely close to $1$ the strategic individual needs to either do nothing (the goal is already achieved), apply $\Theta(\sqrt n)$ vote operations, apply $\Theta(n)$ vote operations, or  the goal cannot be achieve no matter how many vote operations are applied. This characterization holds for any integer generalized scoring rule, any set of vote operations, and any distribution $\pi$ for individual votes.

The proof of Theorem~\ref{thm:main} is based on the Central Limit Theorem and on sensitivity analyses for the integer linear programmings (ILPs). It works as follows. We will formulate each of the strategic individual's three goals as a set of ILPs in Section~\ref{sec:ilp}. By applying Central Limit Theorem, we show that with probability that goes to $1$ the random generated preference-profile satisfies a desired property. Then, for each such preference-profile we apply the sensitivity analyses in~\cite{Cook86:Sensitivity}  to show that with high probability the number of operations that are necessary is either $0$, $\Theta(\sqrt n)$, $\Theta(n)$, or $\infty$.

While Theorem~\ref{thm:main} may look quite abstract, we show later in the paper that many well-studied types of strategic behavior fall under our vote operation framework, including constructive/destructive manipulation, bribery, and control by adding/deleting votes, margin of victory, and minimum manipulation coalition size.\footnote{We defer the definition of these types of strategic behavior to Section~\ref{sec:application}.} Therefore, we naturally obtain corollaries of Theorem~\ref{thm:main} for these types of strategic behavior. The theorem also applies to other types of strategic behavior, for example the mixture of any types mentioned above, which is known as {\em multimode control attacks}~\cite{Faliszewski11:Multimode}.

\subsection{Related Work and Discussion}
Our main theorem applies to any integer generalized scoring rule for destructive manipulation, constructive and destructive  bribery and control by adding/deleting votes. To the best of our knowledge, no similar results were obtained even for specific voting rules for these types of strategic behavior.
Three previous papers obtained similar results  for manipulation, margin of victory, and minimum manipulation coalition size. The applications of our main theorem to these types of strategic behavior are slightly weaker, but we stress that our main theorem is significantly more general. 

\vspace{2mm}
\noindent{\bf Three related papers.} 
First, the dichotomy theorem in~\cite{Xia08:Generalized} 
implies that, (informally) when the votes are drawn i.i.d.~from {\em some} distribution, with probability that goes to $1$ the solution to constructive and destructive {\sc UCO} is either $0$ or approximately $\sqrt n$ for {\em some} favored alternatives. However, this result only works for the {\sc UCO} problem and some distributions over the votes.

Second, it was proved in~\cite{Xia12:Computing} that for any non-redundant generalized scoring rules that satisfy a continuity condition, when the votes are drawn i.i.d.~and we let the number of voters $n$ go to infinity, either with probability that can be arbitrarily close to $1$ the margin of victory is $\Theta(\sqrt n)$, or with probability that can be arbitrarily close to $1$ the margin of victory is $\Theta(n)$. It is easy to show that for non-redundant voting rules, the margin of victory is never $0$ or $\infty$. Though it was shown in~\cite{Xia12:Computing} that many commonly studied voting rules are GSRs that satisfy such continuity condition, in general it is not clear how restrictive the continuity condition is. More importantly, the result only works for the margin of victory problem.

Third, in~\cite{Pritchard09:Asymptotics}, the authors investigated the distribution over the minimum manipulation coalition size for positional scoring rules when the votes are drawn i.i.d.~from the uniform distribution. However, it is not clear how their techniques can be extended beyond the uniform distributions and positional scoring rules, which are a very special case of generalized scoring rules. Moreover, the paper only focused on the  minimum manipulation coalition size problem.

Our results has both negative and positive implications. On the negative side, our results provide yet another evidence that computational complexity is not a strong barrier against strategic behavior, because the strategic individual now has some information about the number of operations that are needed, without spending any computational cost or even without looking at the input instance. Although the estimation of our theorem may not be very precise (because we do not know which of the four cases a given instance belongs to), such estimation may be explored to designing effective algorithms that facilitate strategic behavior. On the positive side, this easiness of computation is not always a bad thing: sometimes we want to do such computation in order to test how robust a given preference-profile is. For example, computing the margin of victory is an important component in designing novel {\em risk-limiting audit methods}~\cite{Magrino11:Computing,Cary11:Estimating,Xia12:Computing,Stark08:Conservative,Stark08:Sharper,Stark09:Risk,Stark09:Efficient,Stark10:Super}.

While being quite general, our results have  two main limitations. First, they are asymptotical results, where we fix the number of alternatives and let the number of voters go to infinity. We do not know the convergence rate, or equivalently, how many voters are needed for the observation to hold. In fact, this is a standard setting  in previous work, especially in the studies of ``frequency of manipulability''. We feel that our results work well in settings where there are small number of alternatives and large number of voters, e.g., political elections. Second, our results show that with high probability one of the four cases holds ($0$, $\Theta(\sqrt n)$, $\Theta(n)$, $\infty$), but we do not know which case holds more often. We will briefly discuss this issue in~\ref{app:discussion}. It is possible to refine our study for specific voting rules and specific types of strategic behavior that fall under our framework, which we leave as future work.


\subsection{Structure of the Paper}

After recalling basic definitions of voting and generalized scoring rules, we present the framework in Section~\ref{sec:model}, where we define vote operations as well as the strategic individual's objectives. Then, in Section~\ref{sec:ilp} we formulate the optimization problem for the strategic individual as a set of integer linear programmings (ILPs). The main theorem will be presented in Section~\ref{sec:main}. To show the wide application of the framework and the main theorem, we show that many commonly studied types of strategic behavior can be modeled as vote operations for generalized scoring rules in Section~\ref{sec:application}, which means that our main theorem naturally applies to these cases. We add some discussions and point out some future directions in Section~\ref{sec:future} and~\ref{app:discussion}. Some proofs are relegated to~\ref{app:proofs}.

\section{Preliminaries}\label{sec:prelim}

Let $\mc$ denote the set of {\em alternatives} (or {\em candidates}), $|\mc|=m$. We assume strict preference orders.  That is, a vote is a
linear order over $\mc$.  The set
of all linear orders over $\mc$ is denoted by $L(\mc)$.  A {\em
  preference-profile} $P$ is a collection of $n$ votes for some $n\in
\mathbb N$, that is, $P\in L(\mc)^n$. Let $L(\mc)^*=\bigcup_{n=1}^\infty L(\mc)^n$. A {\em voting rule} $r$ is a
mapping that assigns to each preference-profile a single winner. That is, $r: L(\mc)^*\ra
\mc$. Throughout the paper, we let $n$ denote
the number of votes and let $m$ denote the number of alternatives.

We now recall the definition of {\em generalized scoring rules (GSRs)}~\cite{Xia08:Generalized}. For any $K\in\mathbb N$, let $\mo_K=\{o_1,\ldots,o_K\}$. A {\em total preorder} ({\em preorder} for short) is a reflexive, transitive, and total relation. Let $\text{Pre}(\mo_K)$ denote the set of all preorders over $\mo_K$. For any $\vec p\in {\mathbb R}^K$, we let $\order(\vec p)$ denote the preorder $\prefe$ over $\mo_K$ where $o_{k_1}\prefe o_{k_2}$ if and only if $p_{k_1}\geq p_{k_2}$. That is,
the $k_1$-th component of $\vec p$ is as large as the $k_2$-th component of $\vec p$. For any preorder $\prefe$, if $o\prefe o'$ and $o'\prefe o$, then we write $o=_\prefe o'$. Each preorder $\prefe$ naturally induces a (partial) strict order $\pref$, where $o\pref o'$ if and only if $o\prefe o'$ and $o'\ntrianglerighteq o$. 

\begin{dfn}
  Let $K\in \mathbb N$, $f:L(\mc)\ra {\mathbb R}^K$ and $g:\text{Pre}(\mo_K)\ra
  \mc$.  $f$ and $g$ determine a
  {\em
    generalized scoring rule (GSR)} $GS(f,g)$ as follows. 
  For any preference-profile
  $P=(V_1,\ldots, V_n)\in L(\mc)^n$, abusing the notation we let $f(P)=\sum_{i=1}^nf(V_i)$, and 
let $GS(f,g)(P)=g(\order(f(P)))$. 
We say that
$GS(f,g)$ is {\em of order $K$}.
If $f(V)\in {\mathbb Z}^K$ holds for all $V\in L(\mc)$, then we call $\gs(f,g)$ an {\em integer GSR}.
\end{dfn} 
For any $V\in L(\mc)$, $f(V)$ is called a {\em generalized scoring vector}, $f(P)$ is called a {\em total generalized scoring vector}, and $\order(f(P))$ is called the {\em induced preorder} of $P$.  The class of integer GSRs is equivalent to the class of {\em rational} GSRs, where the components of each generalized scoring vector is in $\mathbb Q$, because for any $l>0$, $\gs(f,g)=\gs(l\cdot f,g)$.

Almost all commonly studied voting rules are generalized scoring rules, including (but not limited to) approval voting, Bucklin, Copeland, maximin, plurality with runoff, ranked pairs, and multi-stage voting rules that use GSRs in each stage to eliminate alternatives (including Nanson's and Baldwin's rule). 
As an example, we recall the proof from~\cite{Xia08:Generalized} that the {\em single transferable vote (STV)} rule (a.k.a.~{\em instant-runoff voting} or {\em alternative vote} for single-winner elections) is an integer generalized scoring rule.
\begin{ex}  STV selects the winner in $m$
rounds. In each round, the   alternative that gets the lowest plurality score (the number of times that the alternative is ranked in the top position) drops out, and is removed
  from all of the votes (so that votes for this alternative transfer to
  another alternative in the next round). Ties are broken alphabetically. The last-remaining alternative is the
  winner.
  
 To see that STV is an integer GSR, we will use generalized scoring
  vectors with many components.  For every proper subset $S$ of alternatives,
  for every alternative $c$ outside of $S$, there is a component in the
  vector that contains the number of times that $c$ is ranked first if all
alternatives in $S$ are removed.  Let

\noindent $\bullet$ $K_{STV}=\sum_{i=0}^{m-1}{m \choose i} (m-i)$; the components are
  indexed by $(S,j)$, where $S$ is a proper subset of $\mc$ and $j\leq m,
  c_j \notin S$. 

\noindent $\bullet$ $(f_{STV}(V))_{(S,j)}=1$, if after removing $S$ from
  $V$, $c_j$ is at the top; otherwise, let
  $(f_{STV}(V))_{(S,j)}=0$. 

\noindent $\bullet$ $g_{STV}$ selects the winners based on $\prefe$ as follows. In
the first round, let $j_1$ be the index such that $o_{(\emptyset,j_1)}$ is ranked the lowest in $\prefe$ among all $o_{(\emptyset,j)}$ (if there are multiple such $j$'s, then we break ties alphabetically to select the least-preferred one).  Let
  $S_1=\{c_{j_1}\}$. Then, for any $2\leq i\leq m-1$, define $S_i$
  recursively as follows: $S_i=S_{i-1}\cup \{c_{j_i}\}$, where
  $j_i$ is the index such that $o_{(S_{i-1},j_i)}$ is ranked the lowest in $\prefe$ among all $o_{(S_{i-1},j)}$; finally, the winner is the
  unique alternative in $(\mc\setminus S_{m-1})$.
  \end{ex}
  
GSRs admit a natural axiomatic characterization~\cite{Xia09:Finite}. That is, GSRs are the class of voting rules that satisfy {\em anonymity, homogeneity,} and {\em finite local consistency}. Anonymity says that the winner does not depend on the name of the voters, homogeneity says that if we duplicate the preference-profile multiple times, then the winner does not change, and finite local consistency is an approximation to the well-studied {\em consistency} axiom.  Not all voting rules are GSRs, for example, {\em Dodgson's rule} is not a GSR because it violates homogeneity~\cite{Brandt09:Some}, and the following skewed majority rule is also not a GSR because it also violates homogeneity.

\begin{ex}\label{ex:emajority} For any $\frac{1}{2}<\gamma<1$, the {\em $\gamma$-majority} rule is defined for two alternatives $\{a,b\}$ as follows: $b$ is the winner if and only if the number of voters who prefer $b$ is more than the number of voters who prefer $a$ by at least $n^{\gamma}$.
\end{ex}
Admittedly, these $\gamma$-majority rules are quite artificial. Later in this paper we will see that the observation made for GSRs in our main theorem (Theorem~\ref{thm:main}) does not hold for $\gamma$-majority rules for any $\frac{1}{2}<\gamma<1$. Notice that these rules satisfy anonymity, which means that the observation made in Theorem~\ref{thm:main} cannot be extended to all anonymous voting rules.

\section{Vote Operations} \label{sec:model}
All types of strategic behavior mentioned in the introduction share the following common characteristics.
The strategic individual (who can be  a group of manipulators, a briber, or a controller, etc.) changes the winner by changing the votes in the preference-profile. Therefore, for generalized scoring rules, any such an operation can be  uniquely represented by changes in the total generalized scoring vector. This is in contrast to some other types of strategic behavior where the strategic individual changes the set of alternatives or the voting rule~\cite{Bartholdi92:How,Tideman87:Independence}. 

In this section, we first define the set of operations the strategic individual can apply, then define her goals.
Given a generalized scoring rule of order $K$, we model the strategic behavior, called {\em vote operations}, as a set of vectors, each of which has $K$ elements, representing the changes made to the total generalized scoring vector if the strategic individual applies this operation. We focus on integer vectors in this paper.

\begin{dfn}
Given a GSR $\gs(f,g)$ of order $K$, let $\Delta=[\vec \delta_1 \cdots \vec \delta_T]$ denote the {\em  vote operations}, where for each $i\leq T$, $\vec \delta_i\in {\mathbb Z}^K$ is a column vector that represents the changes made to the generalized scoring vector by applying the $i$-th vote operation. For each  $l\leq K$, let $\Delta_l$ denote the $l$-th row of $\Delta$.\end{dfn}

We will show examples of these vote operations for some well-studied types of strategic behavior  in Section~\ref{sec:application}. 
Given the set of available operations $\Delta$, the strategic individual's behavior is characterized by a vector $\vec v\in {\mathbb N}_{\geq 0}^{T}$, where $\vec v$ is a column vector and for each $i\leq T$, $v_i$ represents the number of $i$-th operation (corresponding  to $\vec \delta_i$) that she applies.  Let 
$\|\vec v\|_1=\sum_{i=1}^Tv_i$ denote the total number of operations in $\vec v$, which is the L$_1$-norm of $\vec v$. It follows that $\Delta\cdot \vec v$ is the change in the total generalized scoring vector introduced by the strategic individual, where for any $l\leq K$, $\Delta_l\cdot \vec v$ is the change in the $l$-th component.

Next, we give definitions of the  strategic individual's three goals and the corresponding computational problems studied in this paper.
\begin{dfn} In the  {\sc constructive vote operation (CVO)} problem, we are given a generalized scoring rule $\gs(f,g)$, a preference-profile $P$, a favored alternative $c$, and a set of vote operations $\Delta=[\vec \delta_1 \cdots \vec \delta_T]$, and we are asked to compute the smallest number $k$, denoted by $\cvo(P,c)$, such that there exists a vector $\vec v\in {\mathbb N}_{\geq 0}^{T}$ with $\|\vec v\|_1= k$ and $g\left(\order(f(P)+\Delta\cdot \vec v)\right)=c$. If such  $\vec v$ does not exist, then we denote $\cvo(P,c)=\infty$.

The {\sc destructive vote operation (DVO)} problem is defined similarly, where $c$ is the disfavored alternative, and we are asked to compute the smallest number $k$, denoted by $\dvo(P,c)$, such that there exists a vector $\vec v\in {\mathbb N}_{\geq 0}^{T}$ with $\|\vec v\|_1= k$ and $g\left(\order(f(P)+\Delta\cdot \vec v)\right)\neq c$.

In the {\sc change-winner vote operation (CWVO)} problem, we are not given $c$ and we are asked to compute $\dvo(P,\gs(f,g)(P))$, denoted by $\cwvo(P)$.
\end{dfn}
In \cvo{}, the strategic individual seeks to make $c$ win; in \dvo{}, the strategic individual seeks to make $c$ lose; and in \cwvo{}, the strategic individual seeks to change the current winner.
 
For a given instance $(P,r)$, $\cwvo$ is a special case of $\dvo$, where $c=\gs(f,g)(P)$. We distinguish these two problems because in this paper, the input preference-profiles are generated randomly, so the winners of these preference-profiles might be different. Therefore, when the preference-profiles are randomly generated, the distribution for the solution to $\dvo$ does not immediately give us a distribution for the solution to $\cwvo$.
  
 \section{The ILP Formulation}\label{sec:ilp}
 Let us first put aside the strategic individual's goal for the moment (i.e., making a favored alternative win, making a disfavored alternative lose, or changing the winner) and focus on the following question: given a preference-profile $P$ and a preorder $\prefe$ over the $K$ components of the generalized scoring vector, that is, $\prefe\in \text{Pre}(\mo_K)$, how many vote manipulations are needed to change the order of the total generalized scoring vector to $\prefe$? Formally, given a $\gs(f,g)$, a preference-profile $P$ and $\prefe\in \text{Pre}(\mo_K)$, we are interested in $\min \{\|\vec v\|_1: \vec v\in {\mathbb N}_{\geq 0}^K, \order(f(P)+\Delta\cdot\vec v)=\prefe\} $.

This can be computed by the following integer linear programming ILP$_\prefe$, where $v_i$ represents the $i$th component in $\vec v$, which must be a nonnegative integer. We recall that $\Delta_l$ denotes the $l$-th row vector of $\Delta$.

\begin{align*}
\begin{tabular}{r@{\hspace{1cm}}rl}
& $\min$&$\|\vec v\|_1$
\\
s.t. &
$\forall o_i=_\prefe o_j:$ & $(\Delta_i-\Delta_j)\cdot \vec v= [f(P)]_j-[f(P)]_i$\\ 
&
$\forall o_i\pref o_j:$ & $(\Delta_i-\Delta_j)\cdot \vec v\geq  [f(P)]_j-[f(P)]_i+1$\\ 
&$\forall i:$& $v_i\geq 0$
\end{tabular}&
&(\text{LP}_{\prefe})
\end{align*}
Now, we take the strategic individual's goal into account. 
We immediately have the following lemma as a warmup, whose proofs are straightforward and are thus omitted.
\begin{lemma}\label{lemma:finite}
Given a GSR $\gs(f,g)$, an alternative $c$, and a preference-profile $P$, 

$\bullet$ $\cvo(P,c)<\infty$ if and only if there exists  $\prefe$ such that $g(\prefe)=c$ and LP$_\prefe$ has an integer solution; 

$\bullet$ $\dvo(P,c)<\infty$ if and only if there exists $\prefe$ such that $g(\prefe)\neq c$ and LP$_\prefe$ has an integer solution;

$\bullet$ $\cwvo(P)<\infty$ if and only if there exists  $\prefe$ such that $g(\prefe)\neq \gs(f,g)(P)$ and LP$_\prefe$ has an integer solution. (We do not need the input $c$ for this problem.)
\end{lemma}
Moreover, the solution to each of the three problems is the minimum objective value in all LPs corresponding to the problem. For example, if $\cvo(P,c)<\infty$, then 

$\hfill \cvo(P,c)=\min_{\|\vec v\|_1}\{\vec v\text{ is the solution to some LP}_\prefe\text{ where }g(\prefe)=c\}\hfill $

\section{The Main Theorem}\label{sec:main}
In this section we prove the main theorem, which states that for any fixed $m$, any generalized scoring rules, and any set of vote operations $\Delta$, if $n$ votes are generated i.i.d., then for \cvo{} (respectively, \dvo{}, \cwvo{}), with probability that can be infinitely close to $1$, the solution is either $0$, $\Theta(\sqrt n)$, $\Theta(n)$, or $\infty$.

We first present a simple example for the majority rule for two alternatives $\{a,b\}$ to show the taste of the proof for a very special case, and then comment on why this idea cannot be extended to GSRs. After the proof of Theorem~\ref{thm:main} we will add more comments on the non-triviality of proof.

\begin{ex}\label{ex:motivation}Suppose there are $n$ voters, whose votes are drawn i.i.d.~from a distribution $\pi$ over all possible votes (i.e., voting for $a$ with probability $\pi(a)$ and voting for $b$ with probability $\pi(b)$, w.l.o.g.~$\pi(a)\geq \pi(b)$). Let $Y_a$ (respectively, $Y_b$) denote random variable that represents the total number of voters for $a$  (respectively, for $b$). The number of manipulators that are needed to make $b$ to win (i.e., the solution to the UCO problem, see Section~\ref{subsec:uco} for formal definition) is thus a random variable $Y_a-Y_b$.\footnote{If $Y_a-Y_b<0$ then no manipulator is needed.} Let $X$ denote the random variable that takes $1$ with probability $\pi(a)$ and takes $-1$ with probability $\pi(b)$. It follows that $Y_a-Y_b=\underbrace{X+\cdots+X}_{n}$. By the Central Limit Theorem, $Y_a-Y_b$ converges to a normal distribution with mean $n\cdot E(X)$ and variance ${n\cdot \text{Var}(X)}$.

We are interested in usually how large is $Y_a-Y_b$. Not surprisingly, the answer depends on the distribution $\pi$. If $\pi(a)=\pi(b)=1/2$, then the mean of $Y_a-Y_b$ is zero, and the probability that it is a few standard deviations away from the mean is small. For example, the probability that its absolute value is larger than $4\sqrt{n\cdot \text{Var}(X)}$ is less than $0.01$, which means that with $99\%$ probability the solution of UCO is no more than $4\sqrt{n\cdot \text{Var}(X)}$. On the other hand, if $\pi(a)> \pi(b)$, then the mean of $Y_a-Y_b$ is $n(\pi(a)-\pi(b))$, which means that with high probability the solution of UCO is very close to $n(\pi(a)-\pi(b))=\Theta(n)$.
\end{ex}
The idea behind the argument in Example~\ref{ex:motivation} can be easily extended to positional scoring rules~\cite{Procaccia07:Average}. However, we do not believe that it can be extended to generalized scoring rules, even for the case of manipulation, for the following two reasons. First, for generalized scoring rules, the components of the generalized scoring vector do not correspond to the ``scores'' of alternatives. Therefore, having two components tied in the total generalized scoring vector does not mean that two alternatives are tied. Second, the conditions for an alternative to win can be much more complicated than the condition for positional scoring rules, which amounts to requiring that a corresponding component of the total generalized scoring vector is the largest. Therefore, it is not easy to figure out whether the manipulators can achieve their goal by just knowing the asymptotic relationship between the components of the total generalized scoring vector.

\begin{thm}\label{thm:main}
Let $\gs(f,g)$ be an integer generalized scoring rule, let $\pi$ be a distribution over all linear orders, and let $\Delta$ be a set of  vote operations. Suppose we fix the number of alternatives, generate $n$ votes i.i.d.~according to $\pi$, and let $P_n$ denote the preference-profile. Then, for any alternative $c$, $\vo\in\{\cvo,\dvo,\cwvo\footnote{When $\vo=\cwvo$, we let $\vo(P_n,c)$ denote $\cwvo(P_n)$.}\}$, and any $\epsilon >0$, there exists $\beta^*>1$ such that as $n$ goes to infinity, the total probability for the following four events sum up to more than $1-\epsilon$: 
(1) $\vo(P_n,c)=0$, (2) $\frac{1}{\beta^*}\sqrt n<\vo(P_n,c)<\beta^* \sqrt n$, (3) $\frac{1}{\beta^*} n<\vo(P_n,c)<\beta^* n$, and (4) $\vo(P_n,c)=\infty$.
\end{thm}
\begin{prf}{thm:main} Let 
$f(P_\pi)=\sum_{V\in L(\mc)}\pi(V)\cdot f(V)$, and $\prefe_\pi=\order(f(P_\pi))$.  We first prove the theorem for $\cvo$, and then show how to adjust the proof for $\dvo$ and $\cwvo$. The theorem is proved in the following two steps. Step 1: we show that as $n$ goes to infinity, with probability that goes to one we have the following: in a randomly generated $P_n$, the difference between any pair of components in $f(P_n)$ is either $\Theta(\sqrt n)$ or $\Theta(n)$. Step 2: we apply sensitivity analyses to ILPs that are similar to the ILP given in Section~\ref{sec:ilp}  to prove that for any such preference-profile and any $\vo\in\{\cvo,\dvo,\cwvo\}$, $\vo(P_n,c)$ is either $0$, $\Theta(\sqrt n)$, $\Theta(n)$, or $\infty$. The idea behind Step 2 is, for any preference-profile $P_n$, if the difference between a pair of components in $f(P_n)$ is $\Theta(\sqrt n)$, then we consider this  pair of components (not alternatives) to be ``almost tied''; if the difference is $\Theta(n)$, then we consider them to be ``far away''. Take $\cvo$ as an example, we can easily identify the cases where $\cvo(P_n,c)$ is either $0$ (when $\gs(f,g)=c$) or $\infty$ (by Lemma~\ref{lemma:finite}). Then, we will first try to break these ``almost tied'' pairs by using LPs that are similar to LP$_\prefe$ introduced in Section~\ref{sec:ilp}, and show that if there exists an integer solution $\vec v$, then the objective value $\|\vec v\|_1$ is $\Theta(\sqrt n)$. Otherwise, we have to change the orders between some ``far away'' pairs by using LP$_\prefe$'s, and show that if there exists an integer solution to some LP$_\prefe$ with $g(\prefe)=c$, then the objective value is $\Theta(n)$.

Formally, given $n\in \mathbb N$ and $\beta>1$, let $\mcp_\beta$ denote the set of all $n$-vote preference-profiles $P$ that satisfy the following two conditions (we recall that  $f(P_\pi)=\sum_{V\in L(\mc)}\pi(V)\cdot f(V)$): for any pair  $i,j\leq K$,

1. if $[f(P_\pi)]_i=[f(P_\pi)]_j$ then $\frac{1}{\beta}\sqrt n<|[f(P)]_i-[f(P)]_j|<\beta\sqrt n$;

2.  if $[f(P_\pi)]_i\neq [f(P_\pi)]_j$ then $\frac{1}{\beta} n<|[f(P)]_i-[f(P)]_j|<\beta n$.

The following lemma was proved in~\cite{Xia12:Computing}, which follows after the Central Limit Theorem. 

\begin{lemma}\label{lemma:limit} For any $\epsilon>0$, there exists $\beta$ such that 
$\lim_{n\ra\infty} \pr{P_n\in \mcp_\beta}>1-\epsilon$.
\end{lemma}

For any given $\epsilon$, in the rest of the proof we fix $\beta$ to be a constant guaranteed by Lemma~\ref{lemma:limit}. The next lemma (whose proof is deferred to the appendix) will be frequently used in the rest of the proof.
\begin{lemma}\label{lemma:sensitivity} Fix an integer matrix $\bf A$. There exists a constant $\beta_{\bf A}$ that only depends on $\bf A$, such that if the following LP has an integer solution, then the solution is no more than $\beta_{\bf A}\cdot \|\vec b\|_\infty$. 
$$\min \|\vec x\|_1, \text{ s.t. }{\bf A}\cdot \vec x\geq \vec b$$
\end{lemma}

To prove that with high probability $\cvo(P_n,c)$ is either $0$, $\Theta(\sqrt n)$, $\Theta(n)$, or $\infty$, we introduce the following notation.  A preorder $\prefe'$ is a {\em refinement} of another preorder $\prefe$, if $\pref'$ extends $\pref$. That is, $\pref\subseteq \pref'$. We note that $\prefe$ is a refinement of itself. 
Let $\prefe'\ominus\prefe$ denote the strict orders that are in $\pref'$ but not in $\pref$. That is, $(o_i,o_j)\in (\prefe'\ominus\prefe)$ if and only if $o_i\pref' o_j$ and $o_i=_{\prefe}o_j$. We define the following LP that is similar to LP$_\prefe$ defined in Section~\ref{sec:ilp}, which  will be used to check whether there is a way to break ``almost tied'' pairs of components to make $c$ win. For any preorder $\prefe$ and any of its refinement $\prefe'$, we define LP$_{\prefe'\ominus\prefe}$ as follows.
\begin{align*}
\begin{tabular}{r@{\hspace{1cm}}rl}
& $\min$&$\|\vec v\|_1$
\\
s.t. &
$\forall o_i=_{\prefe'} o_j:$ & $(\Delta_i-\Delta_j)\cdot \vec v= [f(P)]_j-[f(P)]_i$\\ 
&
$\forall (o_i, o_j)\in (\prefe'\ominus\prefe):$ & $(\Delta_i-\Delta_j)\cdot \vec v\geq  [f(P)]_j-[f(P)]_i+1$\\ 
&$\forall i:$& $v_i\geq 0$
\end{tabular}&
&(\text{LP}_{\prefe'\ominus\prefe})
\end{align*}

LP$_{\prefe'\ominus\prefe}$ is defined with a little abuse of notation because some of its constraints depend on $\prefe$ (not only the pairwise comparisons in $(\prefe'\ominus\prefe)$). This will not cause confusion because we will always indicate $\prefe$ in the subscript. We note that there is a constraint in  LP$_{\prefe'\ominus\prefe}$ for each pair of components $o_i,o_j$ with $o_i=_{\prefe}o_j$. Therefore, LP$_{\prefe'\ominus\prefe}$ is used to find a solution that breaks ties in $\prefe$. It follows that LP$_{\prefe'\ominus\prefe}$ has an integer solution $\vec v$ if and only if the strategic individual can make the order between any pairs of $o_i,o_j$ with $o_i=_{\prefe}o_j$ to be the one in $\prefe'$ by applying the $i$-th operation $v_i$ times, and the total number of vote operations is $\|\vec v\|_1$.


The following two claims identify the preference-profiles in $\mcp_\beta$ for which $\cvo$ is $\Theta (\sqrt n)$ and $\Theta (n)$, respectively, whose proofs are deferred to the appendix.
\begin{claim}\label{claim:sqrt} There exists $N\in\mathbb N$ and $\beta'>1$ such that for any $n\geq N$, any $P\in \mcp_\beta$, if (1) $c$ is not the winner for $P$,  and (2) there exists a refinement $\prefe^*$ of $\prefe_\pi=\order(f(P_\pi))$ such that $g(\prefe^*)=c$ and  LP$_{\prefe^*\ominus \prefe_\pi}$ has an integer solution, then $\frac{1}{\beta'}\sqrt n< \cvo(P,c)<\beta'\sqrt n$.
\end{claim}

\begin{claim}\label{claim:n} There exists $\beta'>1$ such that for any $P\in \mcp_\beta$, if (1) $c$ is not the winner for $P$, (2) there does not exist a refinement $\prefe^*$ of $\prefe_\pi=\order(f(P_\pi))$ such that LP$_{\prefe^*\ominus \prefe_\pi}$ has an integer solution, and (3) there exists $\prefe$ such that $g(\prefe)=c$ and LP$_{\prefe}$ has an integer solution, then  $\frac{1}{\beta'} n< \cvo(P,c)<\beta' n$.
\end{claim}

Lastly, for any $P\in \mcp_\beta$ such that $\gs(f,g)(P)\neq c$, the only case not covered by Claim~\ref{claim:sqrt} and Claim~\ref{claim:n} is that there does not exist $\prefe$ with $\gs(f,g)(\prefe)=c$ such that  LP$_\prefe$ has an integer solution. It follows from Lemma~\ref{lemma:finite} that in this case $\cvo(P,c)=\infty$. We note that $\beta'$ in Claim~\ref{claim:sqrt} and Claim~\ref{claim:n} does not depend on $n$. Let $\beta^*$ be an arbitrary number that is larger than the two $\beta'$s. This proves the theorem for $\cvo$.

For $\dvo$, we only need to change $g(\prefe^*)=c$ to $g(\prefe^*)\neq c$ in Claim~\ref{claim:sqrt}, and change $g(\prefe)=c$ to $g(\prefe)\neq c$ in Claim~\ref{claim:n}. For $\cwvo$, $\cwvo(P)$ is never $0$ and we only need to change $g(\prefe^*)=c$ to $g(\prefe^*)\neq \gs(f,g)(P)$ in Claim~\ref{claim:sqrt}, and change $g(\prefe)=c$ to $g(\prefe) \neq \gs(f,g)(P)$ in Claim~\ref{claim:n}. 
\end{prf}

\noindent {\bf More comments on the non-triviality of the proof.} Lemma~\ref{lemma:limit} has been proved in~\cite{Xia12:Computing}, whose intuition is quite straightforward and naturally corresponds to a random walk in multidimensional space. However, we did not find an obvious connection between random walk theory and the observation made in Theorem~\ref{thm:main}. We believe that it is unlikely that an obvious connection exists. One evidence is that the observation made in Theorem~\ref{thm:main} does not hold for some voting rules. For example, consider the $\gamma$-majority rule defined in Example~\ref{ex:emajority}. It is not hard to see that as $n$ goes to infinity, with probability that goes to $1$ we have $\cvo(P_n,b)=\dvo(P_n,a)=\cwvo(P_n)=n^{\gamma}/2$, which is not any of the four cases described in Theorem~\ref{thm:main} if $\frac{1}{2}<\gamma<1$. (This implies that for any $\frac{1}{2}<\gamma<1$, $\gamma$-majority is not a generalized scoring rule, which we already know because they do not satisfy homogeneity.) Therefore, the proof of Theorem~\ref{thm:main} should involve analyses on the specific structure of GSRs. 

The main difficulty in proving Theorem~\ref{thm:main} is, for generalized scoring rules we have to handle the cases where some components of the total generalized scoring vector are equivalent. This only happens with negligible probability for the randomly generated preference-profile $P_n$, but it is not clear how often the strategic individual can make some components equivalent in order to achieve her goal. This is the main reason for us to convert the vote manipulation problem to multiple ILPs and apply Lemma~\ref{lemma:sensitivity} to analyze them.

\section{Applications of the Main Theorem}\label{sec:application} In this section we show how to apply Theorem~\ref{thm:main} to some well-studied types of strategic behavior, including constructive and destructive unweighted coalitional optimization, bribery and control, and margin of victory and minimum manipulation coalition size. In the sequel, we will use each subsection to define these problems and describe how they fit in our vote operation framework, and how Theorem~\ref{thm:main} applies. In the end of the section we present a unified corollary for all these types of strategic behavior.

\subsection{Unweighted Coalitional Optimization}\label{subsec:uco}

\begin{dfn}
In a constructive (respectively, destructive) {\sc unweighted coalitional optimization (UCO)} problem, we are given a voting rule $r$, a preference-profile 
$P^{NM}$ of the non-manipulators, and a (dis)favored alternative $c\in \mc$. We are asked to compute the smallest number of manipulators who can cast votes $P^{M}$ such that $c=r(P^{NM}\cup P^{M})$ (respectively, $c\neq r(P^{NM}\cup P^{M})$).
\end{dfn}

To see how {\sc UCO} fits in the vote operation model, we view the group of manipulators as the strategic individual, and each vote cast by a manipulator is a vote operation. Therefore, the set of operations is exactly the set of all generalized scoring vectors $\{f(V): V\in L(\mc)\}$. To apply Theorem~\ref{thm:main}, for constructive {\sc UCO} we let $\vo=\cvo$ and for destructive {\sc UCO} we let $\vo=\dvo$.

\subsection{Bribery}
In this paper we are interested in the optimization variant of the bribery problem~\cite{Faliszewski09:How}.
\begin{dfn} In a constructive (respectively, destructive) {\sc opt-bribery} problem, we are given a preference-profile $P$ and a (dis)favored alternative $c\in \mc$. 
We are asked to compute the smallest number $k$ such that the strategic individual can change no more than $k$ votes such that $c$ is the winner (respectively, $c$ is not the winner).
\end{dfn}

To see how {\sc opt-bribery} falls under the vote operation framework, we view each action of ``changing a vote'' as a vote operation. Since the strategic individual can only change existing votes in the preference-profile, we define the set of operations to be the difference between the generalized scoring vectors of all votes and the generalized scoring vectors of votes in the support of $\pi$, that is, $\{f(V)-f(W): V,W\in L(\mc)\text{ s.t. }\pi(W)>0\}$. Then, similarly the constructive variant corresponds to $\cvo$ and the destructive variant corresponds to $\dvo$. In both cases Theorem~\ref{thm:main} cannot be directly applied, because in the ILPs we did not limit the total number of each type of vote operations that can be used by the strategic individual. Nevertheless, we can still prove a similar proposition by taking a closer look at the relationship between $\cvo$ ($\dvo$) and {\sc opt-bribery} as follows: For any preference-profile, the solution to $\cvo$ (respectively, $\dvo$) is a lower bound on the solution to constructive (respectively, destructive)  {\sc opt-bribery}, because in $\cvo$ and $\dvo$ there are no constraints on the number of each type of vote operations. We have the following four cases.

1. If the solution to $\cvo$ ($\dvo$) is $0$, then the solution to  constructive (destructive) {\sc opt-bribery} is also $0$. 

2. If the solution to $\cvo$ ($\dvo$) is $\Theta(\sqrt n)$, as $n$ become large enough, with probability that goes to $1$ each type of votes in the support of $\pi$ will appear $\Theta(n)$, which is $>\Theta(\sqrt n)$, times in the randomly generated preference-profile, which means that there are enough votes of each type for the strategic individual to change. 

3. If the solution to $\cvo$ ($\dvo$) is $\Theta(n)$, then the solution to constructive (destructive) {\sc opt-bribery} is either $\Theta(n)$ (when the strategic individual can change {\em all} votes to achieve her goal), or $\infty$.

4. If the solution to $\cvo$ ($\dvo$) is $\infty$, then the solution to constructive (destructive) {\sc opt-bribery} is also $\infty$. 

It follows that the observation made in Theorem~\ref{thm:main} holds for {\sc opt-bribery}.

\subsection{Margin of Victory (MoV)}
\begin{dfn}
Given a voting rule $r$ and a preference-profile $P$, the {\em margin of victory (MoV)} of $P$ is the smallest number $k$ such that the winner can be changed by changing $k$ votes in $P$.  In the {\sc mov} problem, we are given $r$ and $P$, and are asked to compute the margin of victory.\end{dfn}

For a given instance $(P,r)$, {\sc mov} is equivalent to destructive {\sc opt-bribery}, where $c=r(P)$. However, when the input preference-profiles are generated randomly, the winners in these profiles might be different. Therefore, the corollary of Theorem~\ref{thm:main} for  {\sc opt-bribery} does not directly imply a similar corollary for {\sc mov}. This relationship is similar to the relationship between $\dvo$ and $\cwvo$.

Despite this difference, the formulation of {\sc mov} in the vote operation framework is very similar to that of {\sc opt-bribery}: The set of all operations and the argument to apply Theorem~\ref{thm:main} are the same. The only difference is that for {\sc mov}, we obtain the corollary from the $\cwvo$ part of Theorem~\ref{thm:main}, while the corollary for {\sc opt-bribery} is obtained from the $\cvo$ and $\dvo$ parts of Theorem~\ref{thm:main}.

\subsection{Minimum Manipulation Coalition Size (MMCS)}

The {\sc minimum manipulation coalition size (MMCS)} problem is similar to {\sc mov}, except that in {\sc MMCS} the winner must be improved for all voters who change their votes~\cite{Pritchard09:Asymptotics}.
\begin{dfn}
In an {\sc MMCS} problem, we are given a voting rule $r$ and a preference-profile 
$P$. We are asked to compute the smallest number $k$ such that a coalition of $k$ voters can change their votes to change the winner, and all of them prefer the new winner to $r(P)$.
\end{dfn}

Unlike {\sc mov}, {\sc MMCS} falls under the vote operation framework in the following dynamic way. For each preference-profile, suppose $c$ is the current winner. For each adversarial $d\neq c$, we use $\{f(V)-f(W):V,W\in L(\mc)\text{ s.t. }d\succ_W c\text{ and } \pi(W)>0\}$ as the set of operations. That is, we only allow voters who prefer $d$ to $c$ to participate in the manipulative coalition. We also replace each of LP$_{\prefe}$ and LP$_{\prefe^*\ominus\prefe_\pi}$ by multiple LPs, each of which is indexed by a pair of alternatives $(d,c)$ and the constraints are generated by using the corresponding set of operations. Then, the corollary for {\sc MMCS} follows after a similar argument to that of $\cvo$ in Theorem~\ref{thm:main}.

\subsection{Control by Adding/Deleting Votes (CAV/CDV)}
\begin{dfn} In a constructive (respectively, destructive) {\sc optimal control by adding votes (opt-CAV)}
problem, we are given a preference-profile $P$, a (dis)favored alternative $c\in \mc$, and a set $N'$ of additional votes. We are asked to compute the smallest number $k$ such that the strategic individual can add $k$ votes in $N'$ such that $c$ is the winner (respectively, $c$ is not the winner). \end{dfn}

For simplicity, we assume that $|N'|=n$ and the votes in $N'$ are drawn i.i.d.~from a distribution $\pi'$. To show how {\sc opt-CAV} falls under the vote operation model, we let the set of operations to be the generalized scoring vectors of all votes that are in the support of $\pi'$, that is, $\{f(V): V\in L(\mc)\text{ and }\pi'(V)>0\}$. Then, the corollary follows from the $\cvo$ and $\dvo$ parts of Theorem~\ref{thm:main} via a similar argument to the argument for {\sc opt-bribery}.

\begin{dfn}  In a constructive (respectively, destructive) {\sc optimal control by deleting votes (opt-CDV)}
problem, we are given a preference-profile $P$ and a (dis)favored alternative $c\in \mc$. We are asked to compute the smallest number $k$ such that the strategic individual can delete $k$ votes in $P$ such that $c$ is the winner (respectively, $c$ is not the winner). 
\end{dfn}
To show how {\sc opt-CDV} falls under the vote operation framework, we let the set of operations to be the negation of generalized scoring vectors of votes in the support of $\pi'$, that is, $\{-f(V): V\in L(\mc)\text{ and }\pi'(V)>0\}$. Then, the corollary follows from the $\cvo$ and $\dvo$ parts of Theorem~\ref{thm:main} via a similar argument to the argument for {\sc opt-bribery}.
\subsection{A Unified Corollary}
The next corollary of Theorem~\ref{thm:main} summarizes the results obtained for all types of strategic behavior studied in this section.
\begin{coro}\label{coro:coro:unified} For any integer generalized scoring rule, any distribution $\pi$ over votes, and any $X\in$ $\big($\{constructive, destructive\}$\times$\{{\sc UCO, opt-bribery, opt-CAV, opt-CDV}\}$\big)$$\cup$\{{\sc MoV, MMCS}\},
suppose the input preference-profiles are generated i.i.d.~from $\pi$.\footnote{For {\sc CAV}, the distribution over the new votes can be generated i.i.d.~from a different distribution $\pi'$.} Then, for any alternative $c$ and any $\epsilon >0$, there exists $\beta^*>1$ such that the total probability for the solution to $X$ to be one of the following four cases is more than $1-\epsilon$  as $n$ goes to infinity: 
(1) $0$, (2) between $\frac{1}{\beta^*}\sqrt n$ and $\beta^* \sqrt n$, (3) between $\frac{1}{\beta^*} n$ and $\beta^* n$, and (4) $\infty$.
\end{coro}

\section{Discussions and Future Work}\label{sec:future}
In this paper, we proposed a general framework to study vote operations for generalized scoring rules. Our main theorem is a characterization for the number of vote operations that are needed to achieve the strategic individual's goal. We showed that the main theorem can be applied to many types of strategic behavior and many commonly used voting rules, for most of which no similar results were previously known. We further discuss the generality of our framework in the next two paragraphs.

\vspace{2mm}
\noindent{\bf GSRs vs.~integer GSRs.} Though integer GSRs are a subclass of GSRs, we feel that from a computational point of view, focusing integer generalized scoring rules does not sacrifice much generality. 
Because the $g$ function only depends on the {preorder} among components in the total generalized scoring vector, if the $f$ function is scaled up by a constant, then the $g$ function will select the same winner. Therefore, integer GSRs are equivalent to GSRs where components in the generalized scoring vectors are rational numbers. When the components are irrational numbers, two computational problems arise. First, it is not clear how these irrational numbers are represented, and second, it is hard to compare two irrational numbers computationally, thus even harder to compute the preorder of the components of the total generalized scoring vector. On the other hand, integer GSRs do not have such computational constraints. In fact, all commonly studied voting rules that are known to be GSRs are integer GSRs. Therefore, we believe that our main theorem has a wide application (at least can be applied to many commonly studied voting rules).

\vspace{2mm}
\noindent{\bf On the generality of vote operations.} While the framework we proposed covers many types of strategic behavior, some other types of strategic behavior that have been widely studied are not covered by our framework. These types of strategic behavior can be roughly categorized as follows: (1) controls that changes the set of alternatives, for example, control by adding/deleting alternatives~\cite{Bartholdi92:How} and control by introducing clones of alternatives~\cite{Tideman87:Independence,Elkind10:Cloning}; and (2) controls that change the procedure of voting, for example control by (runoff) partition of alternatives and control by partition of voters~\cite{Bartholdi92:How}, and control by changing the agenda of voting~\cite{Lang09:Sequential}. 
Building a more general framework that covers more types of strategic behavior and studying their properties are interesting directions for future research.

As we discussed in the introduction, on the positive side, our main theorem suggests that computing the margin of victory is usually not hard, which helps implementing efficient post-election auditing methods. One promising future direction is to design practical computational techniques for computing the margin of victory for generalized scoring rules, based on the ILP proposed in this paper. On the negative side, our main theorem suggests that computational complexity might merely be a weak barrier against many types of strategic behavior. Therefore, we should look for new ways to protect voting, for example introducing randomization~\cite{Conitzer03:Universal,Elkind05:Hybrid,Walsh12:Lot,Obraztsova11:Complexity,Obraztsova11:Ties}, using multiple rounds~\cite{Davies12:Complexity,Narodytska12:Combining,Davies12:Eliminating}, or limiting the strategic individuals' information about other voters~\cite{Conitzer11:Dominating}. Another interesting research direction is to investigate the phase transition of the probability for a coalition of strategic individuals to achieve their goal by using vote operations, as it was done for manipulation~\cite{Walsh09:Where,Mossel12:Smooth}.



\appendix
\section{Proofs}\label{app:proofs}

\noindent{\bf Lemma~\ref{lemma:sensitivity}}\   {\em Fix an integer matrix $\bf A$. There exists a constant $\beta_{\bf A}$ that only depends on $\bf A$, such that if the following LP has an integer solution, then the solution is no more than $\beta_{\bf A}\cdot \|\vec b\|_\infty$. 
\begin{equation}\min \|\vec x\|_1, \text{ s.t. }{\bf A}\cdot \vec x\geq \vec b\label{equ:generalLP}\end{equation}
}
\begin{prf}[l]{lemma:sensitivity}
Let $\bf A$ be a $m^*\times n^*$ integer matrix, which includes the constraints $\vec x\geq \vec 0$. Suppose LP~(\ref{equ:generalLP}) has a (non-negative) integer solution. 
We note that $\vec 0$ is an optimal integer solution to 
$\min \vec 1\cdot (\vec x)' \text{ s.t. }{\bf A}\cdot \vec x\geq \vec 0$. Then, it follows from Theorem~5~(ii) in~\cite{Cook86:Sensitivity} that LP~(\ref{equ:generalLP}) has a (non-negative) integer solution $\vec z$ such that 

$$\|\vec z-\vec 0\|_\infty\leq n^*\cdot M({\bf A})\cdot(\|\vec b-\vec 0\|_\infty+2),$$ 
where $M({\bf A})$ is the maximum of the absolute values of the determinants of the square sub-matrices of $\bf A$. Since $\bf A$ is fixed, the right hand side becomes a constant, that is, $\|\vec z\|_\infty=O(\|\vec b\|_\infty)$. Therefore, there exists $\beta_{\bf A}$ such that the optimal value in the ILP~(\ref{equ:generalLP}) is no more than $\vec 1\cdot (\vec z)'\leq n^*\|\vec z\|_\infty\leq \beta_{\bf A}\cdot\|\vec b\|_\infty$.
\end{prf}
\ \\

\noindent{\bf Claim~\ref{claim:sqrt}}\  {\em There exists $N\in\mathbb N$ and $\beta'>1$ such that for any $n\geq N$, any $P\in \mcp_\beta$, if (1) $c$ is not the winner for $P$,  and (2) there exists a refinement $\prefe^*$ of $\prefe_\pi=\order(f(P_\pi))$ such that $g(\prefe^*)=c$ and  LP$_{\prefe^*\ominus \prefe_\pi}$ has an integer solution, then $\frac{1}{\beta'}\sqrt n< \cvo(P,c)<\beta'\sqrt n$.}

\begin{prf}[c]{claim:sqrt} Let $\prefe=\order(f(P))$. Because $\gs(f,g)(P)\neq c$, $g(\prefe)\neq c$. Therefore, the strategic individual has to change the order of some components in the generalized scoring vector to make $c$ win. We note that $P\in\mcp_\beta$, which means that the difference between any pair of components of $f(P)$ is more than $\frac{1}{\beta} \sqrt n$.  Let $d_{max}$ denote the maximum difference between any pair of components in generalized score vectors. That is, $d_{max}=\max_{t,t',L\in L(\mc)}\{(f(L))_t-(f(L))_{t'}\}$.  
In order for $c$ to win, the number of vote operations must be at least $\frac{1}{\beta} \sqrt n/d_{max}$. Therefore, $\cvo(P,c)>\frac{1}{\beta d_{max}} \sqrt n$.

We next show the upper bound. Because $P\in\mcp_\beta$, for any pair $o_i,o_j$ with $o_i=_{\prefe_\pi}o_j$, $|[f(P)]_i-[f(P)]_j|<\beta\sqrt n$. Therefore, the right hand side of each (in)equality in  LP$_{\prefe^*\ominus \prefe_\pi}$ is no more than $\beta \sqrt n$. Applying Lemma~\ref{lemma:sensitivity} to  LP$_{\prefe^*\ominus \prefe_\pi}$, we have that there exists a constant $\beta_{\prefe^*,\prefe_\pi}$ that only depends on $\prefe^*$ and $\prefe_\pi$, and an integer solution $\vec v$ with $\|\vec v\|_1\leq \beta_{\prefe^*,\prefe_\pi}\sqrt n$ (the $\bf A$ matrix in Lemma~\ref{lemma:sensitivity} is fixed because we fix the number of alternatives $m$, and the left hand side of each (in)equality in LP$_{\prefe^*\ominus \prefe_\pi}$ does not depend on $n$).   Let $\beta'$ be the maximum of $d_{max}\beta$ and all $\beta_{\prefe^*,\prefe_\pi}$ (since we fix the number of alternatives, there are finite many $\beta_{\prefe^*,\prefe_\pi}$'s). Since $\beta_{\prefe^*,\prefe_\pi}>\frac{1}{d_{max}\beta}$, $\beta'>1$. It follows that $\|\vec v\|_1<\beta'\sqrt n$. 
We next show that for a sufficiently large $n$, if the strategic individual applies $\vec v$, then the order over components of the total scoring vector will become $\prefe^*$. That is, $c$ can be made win.

The idea is, LP$_{\prefe^*\ominus \prefe_\pi}$ ensures that by applying $\vec v$, ties between the ``almost tied'' components are broken as in $\prefe^*$. Since $\|\vec v\|_1=O(\sqrt n)$, when $n$ is large enough the order between any pair of ``far away'' components will not be affected. Formally, let $\vec x=f(P)+\Delta\cdot \vec v$. That is, $\vec x$ is the total generalized scoring vector after the strategic individual applied $\vec v$. Because $\vec v$ is a solution to LP$_{\prefe^*\ominus \prefe_\pi}$, for any pair $o_i,o_j$ with $o_i=_{\prefe_\pi}o_j$, the order between $o_i$ and $o_j$ in $\prefe^*$ is the same as the order between $o_i$ and $o_j$ in $\order(\vec x)$. Since $\prefe^*$ is an extension of $\prefe_\pi$, if $o_i\pref_\pi o_j$, then we must have $o_i\pref^* o_j$. Therefore, we only need to check that for any $o_i\pref_\pi o_j$, we have $[\vec x]_i>[\vec x]_j$. Because $P\in\mcp_\beta$, $|[f(P)]_i-[f(P)]_j|>\frac{1}{\beta}n$. We note that $\|\vec v\|_1<\beta'\sqrt n$, which means that by applying $\vec v$, the strategic individual can only change the difference between any pair of components by no more than $d_{max} \beta'\sqrt n$. Let $N=(d_{max} \beta'\beta)^2+1$. When $n\geq N$, $d_{max} \beta'\sqrt n<\frac{1}{\beta}n$, which means that for any $o_i\pref_\pi o_j$, applying $\vec v$ will not change the order between $o_i$ and $o_j$ in the total generalized scoring vector. This means that by applying $\vec v$, the strategic individual can make $c$ win. Therefore, $ \cvo(P,c)\leq \|\vec v\|_1<\beta'\sqrt n$. It follows that for any $n\geq N$, $\frac{1}{\beta'}\sqrt n< \cvo(P,c)<\beta'\sqrt n$.
\end{prf}

\ \\

\noindent{\bf Claim~\ref{claim:n}}\  {\em There exists $\beta'>1$ such that for any $P\in \mcp_\beta$, if (1) $c$ is not the winner for $P$, (2) there does not exist a refinement $\prefe^*$ of $\prefe_\pi=\order(f(P_\pi))$ such that LP$_{\prefe^*\ominus \prefe_\pi}$ has an integer solution, and (3) there exists $\prefe$ such that $g(\prefe)=c$ and LP$_{\prefe}$ has an integer solution, then  $\frac{1}{\beta'} n< \cvo(P,c)<\beta' n$.}

\begin{prf}[c]{claim:n} Let $\prefe=\order(f(P))$. Because the premises of Claim~\ref{claim:sqrt} do not hold, the strategic individual has to change the order of some pair of components that are ``far away'' (that is, the difference between them is $\Theta(n)$ before the strategic individual applies vote operations)  to make $c$ win. We note that one operation can only change the difference between a pair of components by at most $d_{max}$. Therefore, $\cvo(P,c)\geq \frac{1}{\beta}n/d_{max}$.

On the other hand, it follows from Lemma~\ref{lemma:finite} and condition (3) in the statement of the claim that $\cvo(P,c)<\infty$. The only thing left to show is that there exists $\beta'>1$ such that $\cvo(P,c)<\beta'n$ for all $P\in\mcp_\beta$ and for all $n$. Because $P\in\mcp_\beta$, for any pair $o_i,o_j$, $|[f(P)]_i-[f(P)]_j|<\beta n$. Applying Lemma~\ref{lemma:sensitivity} to LP$_{\prefe}$ and condition (3) in the statement of the claim, we have that for any $\prefe$, there exits $\beta_{\prefe}$ that only depends on $\prefe$, and an optimal integer solution $\vec v$ such that $\|\vec v\|_1\leq \beta_\prefe n$. Let $\beta'$ be the maximum of $d_{max} \beta$ and all $\beta_{\prefe}$ (again, there are finite number of $\beta_{\prefe}$'s). It follows that $\frac{1}{\beta'} n< \cvo(P,c)<\beta' n$.
\end{prf}

\section{Discussion: How often the solution is $0$ or $\infty$?}\label{app:discussion}
One important question is: how large is the probability that the solution to problems studied in this section is $0$ or $\infty$? Not surprisingly the answer depends on both the voting rule and the type of vote operations. The probability can be large for some voting rules. For example, for any voting rule that always selects a given alternative $d$ as the winner, the solution to $\cvo$ (respectively, $\dvo$) is always $0$ (respectively $\infty$) for $c=d$ and is always $\infty$ (respectively $0$) for $c\neq d$.  However, for common voting rules the alternatives are treated almost equally (except for cases with ties). Therefore, we may expect that for a preference-profile whose votes are generated i.i.d., each alternative has almost the same probability of being selected as the winner. This is indeed the case in all commonly used voting rules, including approval voting, all positional scoring rules (which include Borda, plurality, and veto), plurality with runoff, Bucklin, Copeland, maximin, STV, and ranked pairs. Therefore, for these voting rules, when the votes are drawn i.i.d.~uniformly at random, the probability for $\cvo$ is approximately $\frac{1}{m}$ and the probability for $\dvo$ is approximately $\frac{m-1}{m}$. For $\cwvo$, the answer is never $0$ because changing $0$ votes cannot change the winner.

We would also expect for common voting rules, for some types of strategic behavior studied in this section, with low probability the solution is $\infty$. For {\sc UCO}, the strategic individual can introduce many (but finitely many) votes such that the non-manipulators' votes are negligible. For {\sc opt-bribery} and {\sc mov}, the strategic individual can change all votes to achieve her goal. For {\sc MMCS}, {\sc CAV}, and {\sc CDV}, it is not clear how large such probability is. The following table summarizes folklore results for common voting rules when votes are drawn i.i.d.~uniformly at random.

\begin{table}[H]\centering
\begin{tabular}{|c|c|c|}
\hline \bf Optimal solution is & $\bf 0$&$\boldsymbol\infty$\\
\hline \{Cons.\} $\times$ \{{\sc uco, opt-bribery}\} & $\approx \frac{1}{m}$& $0$\\
\hline \{Des.\} $\times$ \{{\sc uco, opt-bribery}\} & $\approx \frac{m-1}{m}$& $0$\\
\hline  {\sc mov}& $0$& $0$\\
\hline
\end{tabular}
\caption{Probability for solutions to some problems to be $0$ or $\infty$ for common voting rules.}
\end{table}

{\footnotesize

\begin{thebibliography}{53}
\expandafter\ifx\csname natexlab\endcsname\relax\def\natexlab#1{#1}\fi
\expandafter\ifx\csname url\endcsname\relax
  \def\url#1{\texttt{#1}}\fi
\expandafter\ifx\csname urlprefix\endcsname\relax\def\urlprefix{URL }\fi

\bibitem[{Baharad and Neeman(2002)}]{Baharad02:Asymptotic}
Baharad, E., Neeman, Z., 2002. The asymptotic strategyproofness of scoring and
  condorcet consistent rules. Review of Economic Design 4, 331--340.

\bibitem[{Bartholdi et~al.(1989)Bartholdi, Tovey, and
  Trick}]{Bartholdi89:Computational}
Bartholdi, III, J., Tovey, C., Trick, M., 1989. The computational difficulty of
  manipulating an election. Social Choice and Welfare 6~(3), 227--241.

\bibitem[{Bartholdi et~al.(1992)Bartholdi, Tovey, and Trick}]{Bartholdi92:How}
Bartholdi, III, J., Tovey, C., Trick, M., 1992. How hard is it to control an
  election? Math. Comput. Modelling 16~(8-9), 27--40, formal theories of
  politics, II.

\bibitem[{Brandt(2009)}]{Brandt09:Some}
Brandt, F., 2009. Some remarks on {Dodgson's} voting rule. Mathematical Logic
  Quarterly 55, 460--463.

\bibitem[{Cary(2011)}]{Cary11:Estimating}
Cary, D., 2011. Estimating the margin of victory for instant-runoff voting. In:
  Proceedings of 2011 EVT/WOTE Conference.

\bibitem[{Conitzer and Sandholm(2003)}]{Conitzer03:Universal}
Conitzer, V., Sandholm, T., 2003. Universal voting protocol tweaks to make
  manipulation hard. In: Proceedings of the Eighteenth International Joint
  Conference on Artificial Intelligence (IJCAI). Acapulco, Mexico, pp.
  781--788.

\bibitem[{Conitzer et~al.(2011)Conitzer, Walsh, and
  Xia}]{Conitzer11:Dominating}
Conitzer, V., Walsh, T., Xia, L., 2011. Dominating manipulations in voting with
  partial information. In: Proceedings of the National Conference on Artificial
  Intelligence (AAAI). San Francisco, CA, USA, pp. 638--643.

\bibitem[{Cook et~al.(1986)Cook, Gerards, Schrijver, and
  Tardos}]{Cook86:Sensitivity}
Cook, W.~J., Gerards, A. M.~H., Schrijver, A., Tardos, E., 1986. Sensitivity
  theorems in integer linear programming. Mathematical Programming 34~(3),
  251--264.

\bibitem[{Davies et~al.(2012{\natexlab{a}})Davies, Katsirelos, Narodytska,
  Walsh, and Xia}]{Davies12:Complexity}
Davies, J., Katsirelos, G., Narodytska, N., Walsh, T., Xia, L.,
  2012{\natexlab{a}}. {Complexity of and Algorithms for the Manipulation of
  Borda, Nanson and Baldwin's Voting Rules}. Artificial Intelligence, to
  appear.

\bibitem[{Davies et~al.(2012{\natexlab{b}})Davies, Narodytska, and
  Walsh}]{Davies12:Eliminating}
Davies, J., Narodytska, N., Walsh, T., 2012{\natexlab{b}}. {Eliminating the
  Weakest Link: Making Manipulation Intractable?} In: Proceedings of the
  National Conference on Artificial Intelligence (AAAI). Toronto, Canada, pp.
  1333--1339.

\bibitem[{Dobzinski and Procaccia(2008)}]{Dobzinski08:Frequent}
Dobzinski, S., Procaccia, A.~D., 2008. Frequent manipulability of elections:
  The case of two voters. In: Proceedings of the Fourth Workshop on Internet
  and Network Economics (WINE). Shanghai, China, pp. 653--664.

\bibitem[{Dwork et~al.(2001)Dwork, Kumar, Naor, and Sivakumar}]{Dwork01:Rank}
Dwork, C., Kumar, R., Naor, M., Sivakumar, D., 2001. Rank aggregation methods
  for the web. In: Proceedings of the 10th World Wide Web Conference. pp.
  613--622.

\bibitem[{Elkind et~al.(2010)Elkind, Faliszewski, and
  Slinko}]{Elkind10:Cloning}
Elkind, E., Faliszewski, P., Slinko, A., 2010. Cloning in elections. In:
  Proceedings of the National Conference on Artificial Intelligence (AAAI).
  Atlanta, GA, USA, pp. 768--773.

\bibitem[{Elkind and Lipmaa(2005)}]{Elkind05:Hybrid}
Elkind, E., Lipmaa, H., 2005. Hybrid voting protocols and hardness of
  manipulation. In: Annual International Symposium on Algorithms and
  Computation (ISAAC). pp. 24--26.

\bibitem[{Ephrati and Rosenschein(1991)}]{Ephrati91:Clarke}
Ephrati, E., Rosenschein, J.~S., 1991. The {C}larke tax as a consensus
  mechanism among automated agents. In: Proceedings of the National Conference
  on Artificial Intelligence (AAAI). Anaheim, CA, USA, pp. 173--178.

\bibitem[{Faliszewski et~al.(2009)Faliszewski, Hemaspaandra, and
  Hemaspaandra}]{Faliszewski09:How}
Faliszewski, P., Hemaspaandra, E., Hemaspaandra, L.~A., 2009. How hard is
  bribery in elections? Journal of Artificial Intelligence Research 35,
  485--532.

\bibitem[{Faliszewski et~al.(2010)Faliszewski, Hemaspaandra, and
  Hemaspaandra}]{Faliszewski10:Using}
Faliszewski, P., Hemaspaandra, E., Hemaspaandra, L.~A., 2010. Using complexity
  to protect elections. Communications of the ACM 53, 74--82.

\bibitem[{Faliszewski et~al.(2011)Faliszewski, Hemaspaandra, and
  Hemaspaandra}]{Faliszewski11:Multimode}
Faliszewski, P., Hemaspaandra, E., Hemaspaandra, L.~A., 2011. Multimode control
  attacks on elections. Journal of Artificial Intelligence Research 40,
  305--351.

\bibitem[{Faliszewski and Procaccia(2010)}]{Faliszewski10:AI}
Faliszewski, P., Procaccia, A.~D., 2010. {AI}'s war on manipulation: {A}re we
  winning? AI Magazine 31~(4), 53--64.

\bibitem[{Friedgut et~al.(2008)Friedgut, Kalai, and
  Nisan}]{Friedgut08:Elections}
Friedgut, E., Kalai, G., Nisan, N., 2008. Elections can be manipulated often.
  In: Proceedings of the Annual Symposium on Foundations of Computer Science
  (FOCS). pp. 243--249.

\bibitem[{Ghosh et~al.(1999)Ghosh, Mundhe, Hernandez, and Sen}]{Ghosh99:Voting}
Ghosh, S., Mundhe, M., Hernandez, K., Sen, S., 1999. Voting for movies: the
  anatomy of a recommender system. In: Proceedings of the third annual
  conference on Autonomous Agents. pp. 434--435.

\bibitem[{Gibbard(1973)}]{Gibbard73:Manipulation}
Gibbard, A., 1973. Manipulation of voting schemes: {A} general result.
  Econometrica 41, 587--601.

\bibitem[{Isaksson et~al.(2010)Isaksson, Kindler, and
  Mossel}]{Isaksson10:Geometry}
Isaksson, M., Kindler, G., Mossel, E., 2010. {The Geometry of Manipulation: A
  Quantitative Proof of the Gibbard-Satterthwaite Theorem}. In: Proceedings of
  the 51st Annual Symposium on Foundations of Computer Science (FOCS).
  Washington, DC, USA, pp. 319--328.

\bibitem[{Lang and Xia(2009)}]{Lang09:Sequential}
Lang, J., Xia, L., 2009. Sequential composition of voting rules in multi-issue
  domains. Mathematical Social Sciences 57~(3), 304--324.

\bibitem[{Magrino et~al.(2011)Magrino, Rivest, Shen, and
  Wagner}]{Magrino11:Computing}
Magrino, T.~R., Rivest, R.~L., Shen, E., Wagner, D., 2011. {Computing the
  Margin of Victory in IRV Elections}. In: Proceedings of 2011 EVT/WOTE
  Conference.

\bibitem[{Mossel et~al.(2012)Mossel, Procaccia, and Racz}]{Mossel12:Smooth}
Mossel, E., Procaccia, A.~D., Racz, M.~Z., 2012. A smooth transition from
  powerlessness to absolute power.
  \url{http://www.cs.cmu.edu/~arielpro/papers/phase.pdf}.

\bibitem[{Mossel and Racz(2012)}]{Mossel12:quantitative}
Mossel, E., Racz, M.~Z., 2012. A quantitative {Gibbard-Satterthwaite} theorem
  without neutrality. In: Proceedings of the The 44th ACM Symposium on Theory
  of Computing (STOC).

\bibitem[{Narodytska et~al.(2012)Narodytska, Walsh, and
  Xia}]{Narodytska12:Combining}
Narodytska, N., Walsh, T., Xia, L., 2012. Combining voting rules together. In:
  Proceedings of the 20th European Conference on Artificial Intelligence
  (ECAI).

\bibitem[{Obraztsova and Elkind(2011)}]{Obraztsova11:Complexity}
Obraztsova, S., Elkind, E., 2011. On the complexity of voting manipulation
  under randomized tie-breaking. In: Proceedings of the Twenty-Second
  International Joint Conference on Artificial Intelligence (IJCAI). Barcelona,
  Catalonia, Spain, pp. 319--324.

\bibitem[{Obraztsova et~al.(2011)Obraztsova, Elkind, and
  Hazon}]{Obraztsova11:Ties}
Obraztsova, S., Elkind, E., Hazon, N., 2011. Ties matter: {C}omplexity of
  voting manipulation revisited. In: Proceedings of the Tenth International
  Joint Conference on Autonomous Agents and Multi-Agent Systems (AAMAS).
  Taipei, Taiwan, pp. 71--78.

\bibitem[{Peleg(1979)}]{Peleg79:Note}
Peleg, B., 1979. A note on manipulability of large voting schemes. Theory and
  Decision 11, 401--412.

\bibitem[{Pennock et~al.(2000)Pennock, Horvitz, and Giles}]{Pennock00:Social}
Pennock, D.~M., Horvitz, E., Giles, C.~L., 2000. Social choice theory and
  recommender systems: Analysis of the axiomatic foundations of collaborative
  filtering. In: Proceedings of the National Conference on Artificial
  Intelligence (AAAI). Austin, TX, USA, pp. 729--734.

\bibitem[{Pritchard and Wilson(2009)}]{Pritchard09:Asymptotics}
Pritchard, G., Wilson, M.~C., 2009. Asymptotics of the minimum manipulating
  coalition size for positional voting rules under impartial culture behaviour.
  Mathematical Social Sciences 1, 35--57.

\bibitem[{Procaccia and Rosenschein(2007)}]{Procaccia07:Average}
Procaccia, A.~D., Rosenschein, J.~S., 2007. Average-case tractability of
  manipulation in voting via the fraction of manipulators. In: Proceedings of
  the Sixth International Joint Conference on Autonomous Agents and Multi-Agent
  Systems (AAMAS). Honolulu, HI, USA, pp. 718--720.

\bibitem[{Rothe and Schend(2012)}]{Rothe12:Typical}
Rothe, J., Schend, L., 2012. {Typical-Case Challenges to Complexity Shields
  That Are Supposed to Protect Elections Against Manipulation and Control: A
  Survey}. In: International Symposium on Artificial Intelligence and
  Mathematics.

\bibitem[{Satterthwaite(1975)}]{Satterthwaite75:Strategy}
Satterthwaite, M., 1975. Strategy-proofness and {A}rrow's conditions: Existence
  and correspondence theorems for voting procedures and social welfare
  functions. Journal of Economic Theory 10, 187--217.

\bibitem[{Slinko(2002)}]{Slinko02:Asymptotic}
Slinko, A., 2002. On asymptotic strategy-proofness of classical social choice
  rules. Theory and Decision 52, 389--398.

\bibitem[{Slinko(2004)}]{Slinko04:How}
Slinko, A., 2004. How large should a coalition be to manipulate an election?
  Mathematical Social Sciences 47~(3), 289--293.

\bibitem[{Stark(2008{\natexlab{a}})}]{Stark08:Conservative}
Stark, P.~B., 2008{\natexlab{a}}. Conservative statistical post-election
  audits. The Annals of Applied Statistics 2~(2), 550--581.

\bibitem[{Stark(2008{\natexlab{b}})}]{Stark08:Sharper}
Stark, P.~B., 2008{\natexlab{b}}. A sharper discrepancy measure for
  post-election audits. The Annals of Applied Statistics 2~(3), 982--985.

\bibitem[{Stark(2009{\natexlab{a}})}]{Stark09:Efficient}
Stark, P.~B., 2009{\natexlab{a}}. {Efficient post-election audits of multiple
  contests: 2009 California tests}. In: 4th Annual Conference on Empirical
  Legal Studies (CELS).

\bibitem[{Stark(2009{\natexlab{b}})}]{Stark09:Risk}
Stark, P.~B., 2009{\natexlab{b}}. Risk-limiting post-election audits:
  {P}-values from common probability inequalities. IEEE Transactions on
  Information Forensics and Security 4, 1005--1014.

\bibitem[{Stark(2010)}]{Stark10:Super}
Stark, P.~B., 2010. Super-simple simultaneous single-ballot risk-limiting
  audits. In: Proceedings of 2010 EVT/WOTE Conference.

\bibitem[{Tideman(1987)}]{Tideman87:Independence}
Tideman, T.~N., 1987. Independence of clones as a criterion for voting rules.
  Social Choice and Welfare 4~(3), 185--206.

\bibitem[{Walsh(2009)}]{Walsh09:Where}
Walsh, T., 2009. Where are the really hard manipulation problems? {T}he phase
  transition in manipulating the veto rule. In: Proceedings of the Twenty-First
  International Joint Conference on Artificial Intelligence (IJCAI). Pasadena,
  CA, USA, pp. 324--329.

\bibitem[{Walsh and Xia(2012)}]{Walsh12:Lot}
Walsh, T., Xia, L., 2012. Lot-based voting rules. In: Proceedings of the
  Eleventh International Joint Conference on Autonomous Agents and Multi-Agent
  Systems (AAMAS). Valencia, Spain, pp. 603--610.

\bibitem[{Xia(2012)}]{Xia12:Computing}
Xia, L., 2012. Computing the margin of victory for various voting rules. In:
  Proceedings of the ACM Conference on Electronic Commerce (EC). Valencia,
  Spain, pp. 982--999.

\bibitem[{Xia and Conitzer(2008{\natexlab{a}})}]{Xia08:Generalized}
Xia, L., Conitzer, V., 2008{\natexlab{a}}. Generalized scoring rules and the
  frequency of coalitional manipulability. In: Proceedings of the ACM
  Conference on Electronic Commerce (EC). Chicago, IL, USA, pp. 109--118.

\bibitem[{Xia and Conitzer(2008{\natexlab{b}})}]{Xia08:Sufficient}
Xia, L., Conitzer, V., 2008{\natexlab{b}}. A sufficient condition for voting
  rules to be frequently manipulable. In: Proceedings of the ACM Conference on
  Electronic Commerce (EC). Chicago, IL, USA, pp. 99--108.

\bibitem[{Xia and Conitzer(2009)}]{Xia09:Finite}
Xia, L., Conitzer, V., 2009. Finite local consistency characterizes generalized
  scoring rules. In: Proceedings of the Twenty-First International Joint
  Conference on Artificial Intelligence (IJCAI). Pasadena, CA, USA, pp.
  336--341.

\bibitem[{Xia et~al.(2010)Xia, Conitzer, and Procaccia}]{Xia10:Scheduling}
Xia, L., Conitzer, V., Procaccia, A.~D., 2010. A scheduling approach to
  coalitional manipulation. In: Proceedings of the ACM Conference on Electronic
  Commerce (EC). Cambridge, MA, USA, pp. 275--284.

\bibitem[{Zuckerman et~al.(2011)Zuckerman, Lev, and
  Rosenschein}]{Zuckerman11:Algorithm}
Zuckerman, M., Lev, O., Rosenschein, J.~S., 2011. An algorithm for the
  coalitional manipulation problem under maximin. In: Proceedings of the Tenth
  International Joint Conference on Autonomous Agents and Multi-Agent Systems
  (AAMAS). Taipei, Taiwan, pp. 845--852.

\bibitem[{Zuckerman et~al.(2009)Zuckerman, Procaccia, and
  Rosenschein}]{Zuckerman09:Algorithms}
Zuckerman, M., Procaccia, A.~D., Rosenschein, J.~S., 2009. Algorithms for the
  coalitional manipulation problem. Artificial Intelligence 173~(2), 392--412.

\end{thebibliography}

}
\end{document}